\documentclass[runningheads]{llncs}
\usepackage[T1]{fontenc}
\usepackage{graphicx}

\usepackage{hyperref}
\usepackage{booktabs}
\usepackage{multirow}
\usepackage{wrapfig}
\usepackage{amsmath}

\usepackage{color}

\usepackage{cleveref}
\begin{document}
\title{Prevalence calibration as shortcut mitigation}
\author{Mohamed Amine Kina\inst{1} \and
Eike Petersen\inst{2}}
\authorrunning{M. A. Kina and E. Petersen}
\institute{Universit\"at Bremen, Bremen, Germany \and
Fraunhofer Institute for Digital Medicine MEVIS, Bremen, Germany}
\maketitle
\begin{abstract}

Shortcut learning denotes the widespread situation in which a classifier exploits spurious correlations rather than diagnostic features. Existing mitigation strategies mostly aim to learn shortcut-invariant representations; their empirical success is limited and they cannot be applied to classifiers using frozen foundation model encoders.
We propose to reframe shortcut learning as fundamentally a calibration problem: unconstrained learning implicitly calibrates each shortcut group to its training set disease prevalence, rendering the resulting classifier necessarily over-confident in one group and under-confident in the other. Building on this insight, we prevalence-equalize calibration between shortcut groups through two encoder-agnostic methods, an in-processing regularizer and a post-hoc prevalence-equalized recalibration step.
Across chest-drain--pneumothorax benchmarks on CheXpert and SIIM-ACR, spanning fine-tuned CNNs and frozen foundation-model backbones, both methods substantially outperform all baselines.
Post-hoc recalibration of a standard ERM-trained DenseNet raises misaligned-group AUROC from 0.23 to 0.73, indicating that shortcut reliance degrades the classification head rather than the underlying representation.
Besides two new state-of-the-art shortcut mitigation approaches, our findings more fundamentally connect shortcut learning to calibration theory and algorithmic fairness.

\end{abstract}
\section{Introduction}
\begin{wrapfigure}{r}{0.48\textwidth}
    \centering
    \vspace{-0.5cm}  \includegraphics[width=0.5\textwidth]{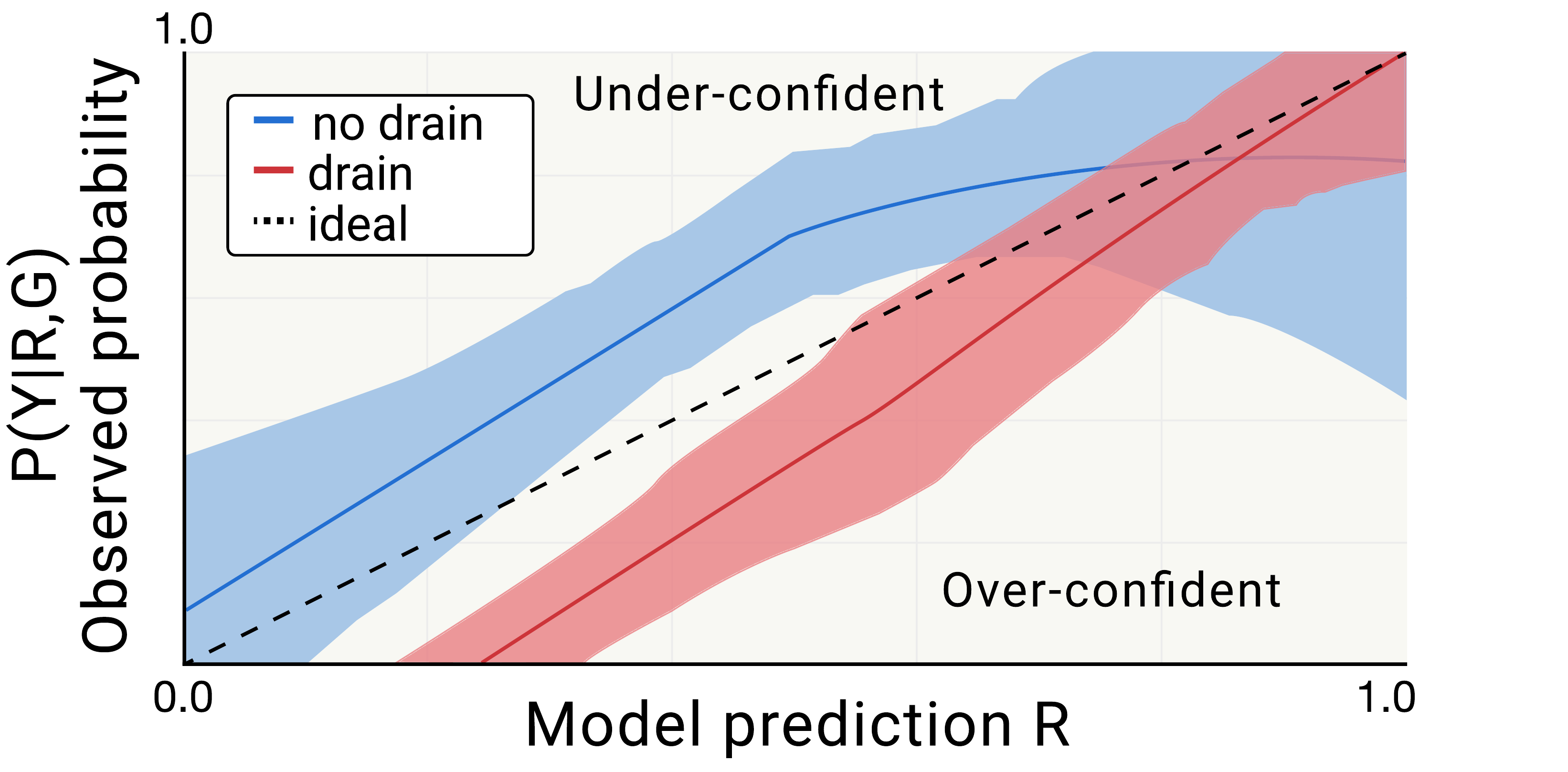}
    \caption{Reliability diagram showing model calibration differences between chest drain subgroups for a MedImageInsight-based pneumothorax classifier on a balanced test set.
    The model exhibits significant miscalibration due to shortcut reliance, presenting as under-confident for the no drain group and over-confident for the drain group.}
    \label{fig:fm-calib-example}
    \vspace{-0.7cm}
\end{wrapfigure}

Shortcut learning represents an essential open problem in medical image classification~\cite{Banerjee2023,Geirhos2020}.
Deep learning models can identify a vast spectrum of non-disease features from medical images and exploit spurious correlations between these features and a given target label~\cite{Geirhos2020,Banerjee2023,Gichoya2022}.
As one often-cited example, models trained to detect pneumothorax in chest x-rays have been shown to rely on the presence of chest drains, a treatment artifact correlated with the disease label rather than a diagnostic feature~\cite{OakdenRayner2020a,Olesen2024,JimenezSanchez2023,Saab2022}. Similar shortcut learning has been documented for hospital-specific markers, patient positioning, and other acquisition-related artifacts, enabling models to achieve high apparent performance while primarily relying on features with no pathophysiological relevance~\cite{Banerjee2023,Geirhos2020}.
Modern pretrained image encoders (foundation models, FMs), while reducing sample size requirements and boosting overall model performance, do not solve the shortcut learning problem: if a non-disease feature can be identified from an image and is correlated with the target label, there is no reason for an FM classification head \emph{not} to rely on these features for improving (apparent) disease classification performance; see \cref{fig:fm-calib-example} for an example from our own study.

Approaches for shortcut learning mitigation have primarily focused on the representation learning stage, aiming for representations that are invariant with respect to the shortcut feature~\cite{Long2018,Yan2020a,Makar2022,Veitch2021,Anthis2023,Feragen2026}.
However, such approaches have fundamental drawbacks.
Due to their reliance on representation-level mitigation, they are incompatible with the increasingly common setting in which classification heads are trained on top of frozen image encoders.
Moreover, they enforce invariances by entirely discarding shortcut-correlated information from the representation, which can degrade downstream target-task performance~\cite{Petersen2023b} and explains limited successes~\cite{Yang2024a,Zong2023}.

Interestingly, as illustrated in \cref{fig:fm-calib-example}, shortcut learning can also be understood as a \emph{calibration} problem: the model has learned to be overconfident in one group and underconfident in the other.
Given that calibration is prevalence-dependent~\cite{Saerens2002} and models tend to optimize for calibration by groups by default~\cite{Liu2019}, this is unsurprising.
Shortcut learning occurs when shortcut features and target labels are strongly correlated, implying large differences in disease prevalence. A well-calibrated model must reflect these differing prevalences by learning distinct disease priors for each group.
Based on this insight, we propose a reframing of shortcut learning mitigation as prevalence-equalized calibration by (shortcut) groups.
Our contributions include:
\begin{enumerate}
    \item a general reframing of shortcut learning as a calibration challenge rather than a problem to be addressed at the level of representation learning,
    \item simple in-processing and post-processing approaches for shortcut learning mitigation based on this insight, which do not require access to or modification of the underlying (pretrained) encoder,
    \item a rigorous hyperparameter optimization scheme specifically tailored for shortcut learning mitigation approaches, addressing the fact that such methods are known to be highly sensitive to regularization strength and enabling optimal tuning of regularization constants and a fair comparison between different mitigation approaches, and
    \item two extensive sets of experiments in chest drain shortcut learning in the well-researched chest x-ray pneumothorax classification setting, showing our proposed mitigation approaches to outperform all baselines by large margins.
\end{enumerate}

\section{Background}
\label{sec:background}

Many shortcut learning mitigation methods have been proposed.
CDAN~\cite{Long2018} and CMMD~\cite{Yan2020a} add a regularization term that penalizes predictability of the shortcut variable from the learned representation, encouraging class-conditional independence between representations and shortcut features~\cite{Petersen2023b}.
JTT~\cite{Liu2021} and G-DRO~\cite{Yang2024a} upweight challenging samples during training to improve worst-group performance.
Deep Feature Reweighting (DFR)~\cite{Kirichenko2023} freezes an ERM-trained feature extractor and retrains only its final classification layer on a small group-balanced held-out set.
Saab et al.~\cite{Saab2022} propose a spatial specificity approach, encouraging models to localize predictions to disease-relevant image regions, yet their approach requires access to bounding box annotations or segmentations.
Empirically, known mitigation strategies achieve only partial success, often trading off overall predictive performance for modest reductions in shortcut reliance, and their effectiveness varies substantially across datasets and shortcut types~\cite{Yang2024a,Zong2023}.

Algorithmic fairness research analyzes notions of group fairness including \emph{equalized odds}~\cite{Hardt2016a},
\emph{balance for the positive and negative class}~\cite{Kleinberg2017,Pleiss2017}, i.e.,
\begin{equation}
    E[R \mid G{=}g_1, Y{=}y] = E[R \mid G{=}g_2, Y{=}y] \quad \forall \, g_1,g_2\in G, y\in Y
    \label{eq:balance}
\end{equation}
with $Y\in\{0,1\}$ denoting the binary prediction target, $R\in[0,1]$ the risk score (confidence) predicted by a model and $G$ group membership (shortcut feature presence or absence),
and \emph{calibration by groups}, i.e.,
\begin{equation}
    P(Y=1 \mid G=g, R=r) = r \quad \forall g\in G, r\in R.
    \label{eq:calibration}
\end{equation}
Liu et al.~\cite{Liu2019} show that \cref{eq:calibration} is a routine consequence of unconstrained learning, whereas other notions such as equalized odds or balance for the positive and negative classes must be explicitly enforced.
Seminal works by Pleiss et al.~\cite{Pleiss2017} and Kleinberg et al.~\cite{Kleinberg2017} prove that in a binary classification setting, satisfying \cref{eq:balance} is incompatible (except for degenerate cases) with satisfying \cref{eq:calibration}, constituting one of the most cited incompatibility results in algorithmic fairness.
Further developing calibration fairness, Hébert-Johnson et al.~\cite{HebertJohnson2018} propose \emph{multicalibration}, extending condition~\eqref{eq:calibration} to all computationally identifiable groups and sparking the fruitful study of omnipredictors~\cite{Gopalan2021}.
Notably, all notions of calibration~\cite{VanCalster2016} imply mean calibration ($E[R]=E[Y]$): the average predicted score must equal the prevalence in the calibration set.
This has important implications for calibration fairness: calibration by groups demands that each group be calibrated to their respective prevalence in the training or calibration dataset; this is the cause of the incompatibility with balance in the case of prevalence differences~\cite{Kleinberg2017,Pleiss2017}.
It also explains why calibration cannot hold under prevalence shifts and models must be recalibrated~\cite{Saerens2002}.
Curiously, the fact that calibration by groups demands groups to be calibrated to their respective \emph{different} prevalences, thus requiring differential group treatment, has escaped attention, so far.

Many have addressed the risks of \emph{demographic} shortcut learning~\cite{Gichoya2022,Banerjee2023,Brown2023}, and others have pointed out that even non-demographic shortcut learning can be a cause of demographic performance disparities~\cite{Olesen2024}.
Causal formulations of shortcut learning and algorithmic fairness often result in a degree of unification of the two fields~\cite{Makar2022,Veitch2021,Anthis2023,Feragen2026}, yet the implications of common notions of group fairness for shortcut learning mitigation have not been explored comprehensively.

\section{Methodology}

\subsection{Datasets}
\label{ssec:data}
\begin{figure}[tb]
    \centering
    \includegraphics[width=\textwidth]{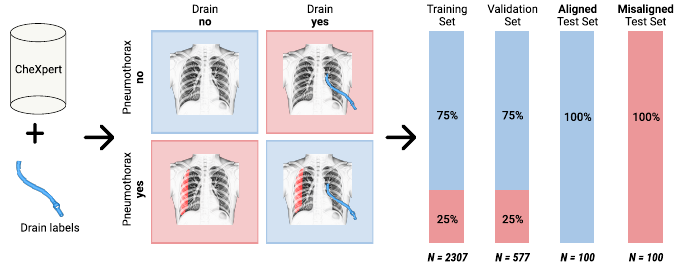}
    \caption{CheXpert data and custom drain labels are categorized into a 2x2 quadrant based on pneumothorax and drain presence. Blue quadrants denote shortcut-aligned cases, while red denote misaligned cases. The resulting data splits consist of shortcut-afflicted training and validation sets and two fully aligned/misaligned test sets.}
    \label{fig:dataset}
\end{figure}

We conduct experiments on chest drain--pneumothorax shortcut learning on CheXpert~\cite{Irvin2019} and the SIIM-ACR Pneumothorax Segmentation dataset.\footnote{\url{https://siim.org/research-journal/siim-machine-learning-challenges/pneumothorax-kaggle-challenge/}}
CheXpert does not natively provide chest-drain annotations.
We obtained drain labels for a subset
from Jiménez-Sanchez et al.~\cite{JimenezSanchez2023} and used these to fine-tune a DenseNet-121
as a drain classifier (AUROC 0.84 on a balanced 60-image test set).
We applied this classifier to all CheXpert frontal radiographs and retained only high-confidence predictions (predicted drain probability lower than the 0.1 quantile of predicted scores on the validation set or greater than the 0.9 quantile of predicted scores on the validation set); the remaining uncertain predictions were discarded.
For constructing the final CheXpert data splits, to minimize the risk of label errors, we follow the methodology of Weng et al.~\cite{Weng2023} and select a single image per patient, prioritizing Pneumothorax-positive and shortcut-misaligned images where available.
For the SIIM-ACR dataset, the pneumothorax target is derived from the segmentation masks (non-empty mask = positive), and we use the binary chest-drain annotations by Saab et al.~\cite{Saab2022}.
We construct shortcut-afflicted training and validation splits with 75\% of pneumothorax-positive cases containing a drain and vice versa.
Test sets were constructed to be fully balanced across all four quadrants;
refer to \cref{fig:dataset} for an illustration.
The final split size is 2307/577/100 and 8460/2115/100 train/val/test for CheXpert and SIIM-ACR, respectively.
All utilized CheXpert samples have drain annotations but not all SIIM-ACR training and validation samples do.

\subsection{Conditional score matching and prevalence-equalized calibration}
As an in-processing shortcut learning mitigation method inspired by the shortcut-learning-as-calibration reframing, we propose Class-Conditional Average Score Matching (CSM).
Specifically, we propose to regularize
\begin{equation}
    \mathcal{L}_{\mathrm{CSM}} = 0.5 \cdot \sigma_G\!\left(E[R \mid G{=}g, Y{=}1]\right) + 0.5 \cdot \sigma_G\!\left(E[R \mid G{=}g, Y{=}0]\right),
    \label{eq:csm}
\end{equation}
that is, the standard deviation $\sigma_G$ of the average predicted model scores across shortcut groups, averaged across the disease-positive and negative strata.
Note that in our case, $g\in\{0,1\}$ as we are considering a single binary shortcut attribute (chest drain presence or absence), but the proposed regularizer trivially generalizes to higher-cardinality settings.
For shortcut mitigation, we optimize
\begin{equation}
    \mathcal{L} = \mathcal{L}_{\mathrm{BCE}} + \lambda_{\mathrm{CSM}} \mathcal{L}_{\mathrm{CSM}},
\end{equation}
where $\mathcal{L}_{\mathrm{BCE}}$ denotes the binary cross-entropy loss and $\lambda_{\mathrm{CSM}}$ a tuning parameter.

The regularizer~\eqref{eq:csm} reaches its global minimum of 0 if the class-conditional average scores coincide between all groups, i.e., if \cref{eq:balance} holds.
As outlined above, this implies an incompatibility with calibration by groups (\cref{eq:calibration}) in the presence of between-group prevalence differences.
However, as also pointed out above, calibration by groups implies calibrating every group to their respective prevalence.
In the context of shortcut learning, this results in a classifier which is overconfident in one shortcut group and underconfident in the other, as illustrated in \cref{fig:fm-calib-example}.
We therefore claim that in the context of shortcut learning, \emph{prevalence-equalized calibration by groups} is the property of interest, i.e.,
\begin{equation}
    P(Y=1 \mid G=g, R=r) = f_{g,\pi^*}(r) \quad \forall g\in G, r\in R,
    \label{eq:prev-calibration}
\end{equation}
where
\begin{equation}
    f_{g,\pi^*}(r) = \frac{\frac{\pi^*}{\pi_g} r}{\frac{\pi^*}{\pi_g} r + \frac{1-\pi^*}{1-\pi_g}(1-r)}
\end{equation}
denotes the formula for recalibrating a risk score $r$ from prevalence $\pi_g=P(y \mid G{=}g)$ to prevalence $\pi^*$~\cite{Saerens2002}.
Note that the choice of $\pi^*\in(0,1)$ is irrelevant; if \cref{eq:prev-calibration} holds for one value of $\pi^*$ it holds for all.
\Cref{eq:prev-calibration} asks that all groups~$g$ be calibrated to the \emph{same} prevalence, regardless of training set prevalence differences.
As a side effect, this resolves the incompatibility of standard calibration by groups with \cref{eq:balance}, since that incompatibility is a result of groups needing to be calibrated to different prevalences~\cite{Kleinberg2017,Pleiss2017}. In fact, \cref{eq:balance} implies $E_{\pi^*}[R \mid G{=}g_1]=E_{\pi^*}[R \mid G{=}g_2] \,\,\forall g_1,g_2,\pi^*$, a necessary condition for \cref{eq:prev-calibration}.

Batchwise estimation of the class-conditional average scores in \cref{eq:csm} requires batches to be well-balanced across shortcut group–-target label combinations.
We compute a (disease-positive or negative) stratum's loss only when all shortcut groups are present in the batch, falling back to plain BCE otherwise. As a more robust alternative, we also implement Dataset Class-Conditional Average Score Matching (DCSM), where we estimate averages across the full training set, storing scores for all training samples while backpropagating only through the current batch.

\subsection{Post-hoc prevalence-equalized shortcut calibration}
The previous section presented an in-processing approach; we also propose a simple post-hoc procedure, simply recalibrating each shortcut group separately to the same reference prevalence~$\pi^*$.
Standard recalibration is trivially achieved using various off-the-shelf calibration schemes; we here employ beta calibration~\cite{Kull2017}.
To adjust for prevalence differences, we recalibrate each group on its share of the validation set using inverse class probabilities as sample weights.
Notice that this results in distinct recalibration mappings for each of the shortcut groups, meaning that at inference time, shortcut group membership must be available.

\subsection{Hyperparameter optimization}
For hyperparameter tuning, we utilize Gaussian process-based Bayesian optimization as implemented in Optuna~\cite{Akiba2019} to maximize the harmonic mean of pairwise validation AUROCs between sets $(Y=1, G=g_i)$, $(Y=0,G=g_j)$, i.e.
\begin{equation}
    \theta^* = \arg\max_\theta \frac{|G|(|G|-1)}{\sum_{g_i \neq g_j} \frac{1}{\text{AUROC}_{ij}}}
    \label{eq:harmonic}
\end{equation}
with $\theta$ the hyperparameters to be optimized.
In the simplest case of a binary-valued shortcut variable, \cref{eq:harmonic} reduces to the harmonic mean of the AUROCs in the shortcut-aligned and the shortcut-misaligned shares of the validation set.
As opposed to the arithmetic mean, the harmonic mean penalizes gains in one group at the other's expense, and as opposed to worst-group performance, it takes all data points into account.
Understanding all three as generalized power means clarifies that the harmonic mean ($p=-1$) interpolates between the other two extremes ($p=1$ for the arithmetic mean, $p=-\infty$ for worst-group performance).

\subsection{Experiments \& evaluation}
We compare the performance of (D)CSM and post-hoc recalibration against baselines including ERM, CDAN, CMMD, JTT, and DFR.
On CheXpert, we evaluate a fully finetuned DenseNet-121 (ImageNet1k-pretrained) as commonly used in prior work, as well as MedImageInsight~\cite{Codella2024} and MedSigLIP~\cite{Sellergren2025} as representative foundation models (FMs); on SIIM-ACR, experiments are limited to the DenseNet-121.
According to their respective documentation, the training set of neither FM included CheXpert or SIIM-ACR.
All models are equipped with a simple 2-layer classification head; foundation model backbones are kept frozen and used as static feature extractors.
Models are trained using AdamW (LR $10^{-4}$ for the classification head and $2 \cdot 10^{-5}$ for the encoder, weight decay 0.005, 150 epochs for the DenseNet and 75 epochs for the FMs), hyperparameters related to the mitigation methods optimized for 50 Optuna trials, best checkpoint selected based on validation performance.
We conduct 10 runs of each experiment.
For the FMs, we evaluate DCSM, post-hoc recalibration, and DFR, all of which operate solely on the classification head and are applicable to frozen encoders.
We evaluate fully shortcut-aligned and misaligned AUROC as well as \cref{eq:harmonic}, which we term \emph{fairness score (FS)}.
Code is available at \url{https://github.com/Amineki6/calibration}.

\section{Results}
\Cref{tab:results-chexpert} presents the results of our experiments.
On CheXpert DenseNet, DCSM and CSM lead the in-processing methods, substantially outperforming the next-best method, CDAN.
Post-hoc recalibration raises standard ERM from FS 0.37 to the overall-best 0.73, exceeding DFR both before (0.56) and after (0.62) recalibration and suggesting that last-layer retraining is unnecessary.
The frozen FMs perform better overall, with further gains from DCSM and recalibrated ERM.
On SIIM-ACR, recalibrated ERM performs best, outperforming the bounding-box-dependent method of Saab et al.~\cite{Saab2022}.

\begin{table}[!t]
  \centering
  \setlength{\abovecaptionskip}{2pt}
  \caption{Aligned/misaligned AUROC and fairness score (mean~$\pm$~std, $N{=}10$). Orig./Recal. denote before/after recalibration; $^*$ denotes BCE selection, otherwise fairness selection. \textbf{Bold}/\underline{underline}/\textit{italics}: top three. CheXpert unless noted.}
  \label{tab:results-chexpert}
  \setlength{\aboverulesep}{0.2ex}
  \setlength{\belowrulesep}{0.3ex}
  \setlength{\cmidrulesep}{1pt}
  \setlength{\tabcolsep}{4pt}
  \newcommand{\pmv}[2]{#1 \ensuremath{\pm} #2}
  \begin{tabular}{cl l ccc}
    \toprule
    & Method & Calib. & Aligned AUROC & Misaligned AUROC & Fairness Score \\
    \midrule
    \multirow{16}{*}{\rotatebox[origin=c]{90}{\textbf{DenseNet}}}
      & \multirow{2}{*}{ERM$^*$}
              & Orig.  & \pmv{0.995}{0.005} & \pmv{0.226}{0.017} & \pmv{0.368}{0.023} \\
      &       & Recal. & \pmv{0.729}{0.034} & \pmv{0.726}{0.033} & \textbf{\pmv{0.727}{0.024}} \\
    \cmidrule(l){2-6}
      & \multirow{2}{*}{ERM}
              & Orig.  & \pmv{0.909}{0.050} & \pmv{0.361}{0.078} & \pmv{0.514}{0.087} \\
      &       & Recal. & \pmv{0.624}{0.105} & \pmv{0.752}{0.049} & \pmv{0.678}{0.078} \\
    \cmidrule(l){2-6}
      & \multirow{2}{*}{CDAN}
              & Orig.  & \pmv{0.691}{0.074} & \pmv{0.675}{0.062} & \pmv{0.679}{0.046} \\
      &       & Recal. & \pmv{0.663}{0.075} & \pmv{0.675}{0.068} & \pmv{0.665}{0.052} \\
    \cmidrule(l){2-6}
      & \multirow{2}{*}{JTT}
              & Orig.  & \pmv{0.670}{0.053} & \pmv{0.616}{0.056} & \pmv{0.639}{0.039} \\
      &       & Recal. & \pmv{0.643}{0.050} & \pmv{0.621}{0.050} & \pmv{0.631}{0.046} \\
    \cmidrule(l){2-6}
      & \multirow{2}{*}{CMMD}
              & Orig.  & \pmv{0.894}{0.027} & \pmv{0.479}{0.047} & \pmv{0.622}{0.040} \\
      &       & Recal. & \pmv{0.669}{0.053} & \pmv{0.694}{0.034} & \pmv{0.681}{0.037} \\
    \cmidrule(l){2-6}
      & \multirow{2}{*}{CSM}
              & Orig.  & \pmv{0.767}{0.057} & \pmv{0.683}{0.043} & \textit{\pmv{0.720}{0.029}} \\
      &       & Recal. & \pmv{0.646}{0.050} & \pmv{0.748}{0.050} & \pmv{0.691}{0.033} \\
    \cmidrule(l){2-6}
      & \multirow{2}{*}{DCSM}
              & Orig.  & \pmv{0.793}{0.057} & \pmv{0.667}{0.051} & \underline{\pmv{0.722}{0.037}} \\
      &       & Recal. & \pmv{0.734}{0.046} & \pmv{0.671}{0.072} & \pmv{0.698}{0.046} \\
    \cmidrule(l){2-6}
      & \multirow{2}{*}{DFR}
              & Orig.  & \pmv{0.806}{0.129} & \pmv{0.452}{0.122} & \pmv{0.563}{0.101} \\
      &       & Recal. & \pmv{0.720}{0.125} & \pmv{0.590}{0.181} & \pmv{0.624}{0.114} \\
    \midrule
    \multirow{8}{*}{\rotatebox[origin=c]{90}{\textbf{MedSigLIP}}}
      & \multirow{2}{*}{ERM$^*$}
              & Orig.  & \pmv{0.893}{0.002} & \pmv{0.568}{0.009} & \pmv{0.694}{0.007} \\
      &       & Recal. & \pmv{0.757}{0.011} & \pmv{0.874}{0.004} & \textbf{\pmv{0.811}{0.006}} \\
    \cmidrule(l){2-6}
      & \multirow{2}{*}{ERM}
              & Orig.  & \pmv{0.878}{0.004} & \pmv{0.577}{0.015} & \pmv{0.696}{0.011} \\
      &       & Recal. & \pmv{0.692}{0.016} & \pmv{0.848}{0.011} & \underline{\pmv{0.762}{0.013}} \\
    \cmidrule(l){2-6}
      & \multirow{2}{*}{DCSM}
              & Orig.  & \pmv{0.681}{0.014} & \pmv{0.857}{0.015} & \textit{\pmv{0.759}{0.009}} \\
      &       & Recal. & \pmv{0.577}{0.022} & \pmv{0.912}{0.008} & \pmv{0.706}{0.018} \\
    \cmidrule(l){2-6}
      & \multirow{2}{*}{DFR}
              & Orig.  & \pmv{0.731}{0.059} & \pmv{0.630}{0.054} & \pmv{0.675}{0.039} \\
      &       & Recal. & \pmv{0.767}{0.045} & \pmv{0.610}{0.066} & \pmv{0.677}{0.046} \\
    \midrule
    \multirow{8}{*}{\rotatebox[origin=c]{90}{\textbf{\shortstack{Med\\ImageInsight}}}}
      & \multirow{2}{*}{ERM$^*$}
              & Orig.  & \pmv{0.898}{0.001} & \pmv{0.642}{0.008} & \pmv{0.749}{0.006} \\
      &       & Recal. & \pmv{0.730}{0.003} & \pmv{0.893}{0.002} & \textit{\pmv{0.803}{0.002}} \\
    \cmidrule(l){2-6}
      & \multirow{2}{*}{ERM}
              & Orig.  & \pmv{0.899}{0.001} & \pmv{0.640}{0.009} & \pmv{0.747}{0.006} \\
      &       & Recal. & \pmv{0.731}{0.003} & \pmv{0.895}{0.001} & \underline{\pmv{0.805}{0.002}} \\
    \cmidrule(l){2-6}
      & \multirow{2}{*}{DCSM}
              & Orig.  & \pmv{0.773}{0.003} & \pmv{0.890}{0.017} & \textbf{\pmv{0.828}{0.007}} \\
      &       & Recal. & \pmv{0.683}{0.014} & \pmv{0.943}{0.002} & \pmv{0.792}{0.009} \\
    \cmidrule(l){2-6}
      & \multirow{2}{*}{DFR}
              & Orig.  & \pmv{0.747}{0.076} & \pmv{0.689}{0.049} & \pmv{0.713}{0.043} \\
      &       & Recal. & \pmv{0.758}{0.063} & \pmv{0.686}{0.044} & \pmv{0.719}{0.044} \\
      \midrule
    \multirow{6}{*}{\rotatebox[origin=c]{90}{\textbf{\shortstack{DenseNet\\(SIIM-ACR)}}}}
      & \multirow{2}{*}{ERM}
              & Orig.  & \pmv{0.991}{0.005} & \pmv{0.496}{0.027} & \pmv{0.660}{0.024} \\
      &       & Recal. & \pmv{0.893}{0.020} & \pmv{0.731}{0.036} & \textbf{\pmv{0.803}{0.025}} \\
    \cmidrule(l){2-6}
      & \multirow{2}{*}{CSM}
              & Orig.  & \pmv{0.926}{0.033} & \pmv{0.584}{0.040} & \pmv{0.715}{0.030} \\
      &       & Recal. & \pmv{0.876}{0.037} & \pmv{0.661}{0.039} & \underline{\pmv{0.753}{0.030}} \\
    \cmidrule(l){2-6}
      & \multirow{2}{*}{DCSM}
              & Orig.  & \pmv{0.873}{0.029} & \pmv{0.613}{0.022} & \pmv{0.719}{0.015} \\
      &       & Recal. & \pmv{0.861}{0.020} & \pmv{0.630}{0.017} & \textit{\pmv{0.727}{0.014}} \\
    \bottomrule
  \end{tabular}
\end{table}

\section{Discussion \& Conclusion}
We reframe shortcut learning as a calibration problem: ERM implicitly calibrates each group to its training prevalence~\cite{Liu2019,Saerens2002}, which differs strongly across shortcut groups.
This motivates our in-processing and post-processing methods, which substantially outperform all baselines across experiments and model classes.

Our empirical findings suggest that shortcut learning need not be treated exclusively as a representation learning problem.
The standard ERM-trained DenseNet obtains AUROC 0.23 on the misaligned test set, indicating that it has fully learned the chest drain-pneumothorax shortcut.
Yet post-hoc prevalence-equalized recalibration raises both aligned and misaligned AUROC to 0.73, showing that diagnostic features remain in the representation and that the classification head is the main bottleneck.
Consistent with DFR's finding that last-layer retraining can recover robustness~\cite{Kirichenko2023}, recalibration alone performs better in our experiments.

Our analyses also have implications for algorithmic fairness: calibration by groups calibrates each group to its own training prevalence, requiring different target priors.
Prevalence-equalized group calibration may therefore be a more desirable aim; future work should explore its implications.

Finally, post-hoc recalibration substantially improves ERM but slightly worsens (D)CSM; explaining this tension remains future work.

\begin{credits}

\subsubsection{\discintname}
The authors declare no competing interests.
\end{credits}
\clearpage
\bibliographystyle{splncs04}
\bibliography{references}

@InProceedings{Saab2022,
  author    = {Saab, Khaled and Hooper, Sarah and Chen, Mayee and Zhang, Michael and Rubin, Daniel and Re, Christopher},
  booktitle = {Proceedings of the 7th Machine Learning for Healthcare Conference},
  title     = {Reducing Reliance on Spurious Features in Medical Image Classification with Spatial Specificity},
  year      = {2022},
  pages     = {760--784},
  publisher = {PMLR},
  series    = {Proceedings of Machine Learning Research},
  volume    = {182},
}

@Article{Yang2024a,
  author    = {Yang, Yuzhe and Zhang, Haoran and Gichoya, Judy W. and Katabi, Dina and Ghassemi, Marzyeh},
  journal   = {Nature Medicine},
  title     = {The limits of fair medical imaging {AI} in real-world generalization},
  year      = {2024},
  issn      = {1546-170X},
  publisher = {Springer Science and Business Media LLC},
}

@InProceedings{Zong2023,
  author    = {Yongshuo Zong and Yongxin Yang and Timothy Hospedales},
  booktitle = {The Eleventh International Conference on Learning Representations},
  title     = {{MEDFAIR}: Benchmarking Fairness for Medical Imaging},
  year      = {2023},
}

@InProceedings{Kirichenko2023,
  author    = {Kirichenko, Polina and Izmailov, Pavel and Wilson, Andrew Gordon},
  booktitle = {The Eleventh International Conference on Learning Representations},
  title     = {Last Layer Re-Training is Sufficient for Robustness to Spurious Correlations},
  year      = {2023},
}

@InProceedings{Olesen2024,
  author    = {Olesen, Vincent and Weng, Nina and Feragen, Aasa and Petersen, Eike},
  pages     = {3--13},
  publisher = {Springer Nature Switzerland},
  title     = {Slicing Through Bias: Explaining Performance Gaps in Medical Image Analysis Using Slice Discovery Methods},
  year      = {2024},
  booktitle = {MICCAI 2024 FAIMI Workshop},
}

@InProceedings{HebertJohnson2018,
  author    = {Hébert-Johnson, {\'U}rsula and Kim, Michael and Reingold, Omer and Rothblum, Guy},
  booktitle = {Proceedings of the 35th International Conference on Machine Learning},
  title     = {Multicalibration: Calibration for the ({C}omputationally-Identifiable) Masses},
  pages     = {1939--1948},
  publisher = {PMLR},
  series    = {Proceedings of Machine Learning Research},
  volume    = {80},
  year      = {2018},
}

@Article{Saerens2002,
  author    = {Saerens, Marco and Latinne, Patrice and Decaestecker, Christine},
  journal   = {Neural Computation},
  title     = {Adjusting the Outputs of a Classifier to New a Priori Probabilities: A Simple Procedure},
  year      = {2002},
  issn      = {1530-888X},
  number    = {1},
  pages     = {21--41},
  volume    = {14},
  publisher = {MIT Press},
}

@InProceedings{Kull2017,
  author    = {Kull, Meelis and Filho, Telmo Silva and Flach, Peter},
  booktitle = {Proceedings of the 20th International Conference on Artificial Intelligence and Statistics},
  year      = {2017},
  title     = {{Beta calibration: a well-founded and easily implemented improvement on logistic calibration for binary classifiers}},
  pages     = {623--631},
  publisher = {PMLR},
  series    = {Proceedings of Machine Learning Research},
  volume    = {54},
}

@Article{Geirhos2020,
  author       = {Robert Geirhos and Jörn-Henrik Jacobsen and Claudio Michaelis and Richard Zemel and Wieland Brendel and Matthias Bethge and Felix A. Wichmann},
  title        = {Shortcut learning in deep neural networks},
  number       = {11},
  pages        = {665--673},
  volume       = {2},
  year         = {2020},
  journal = {Nature Machine Intelligence},
}

@InProceedings{Weng2023,
  author    = {Weng, Nina and Bigdeli, Siavash and Petersen, Eike and Feragen, Aasa},
  booktitle = {MICCAI 2023 FAIMI Workshop},
  title     = {Are Sex-Based Physiological Differences the Cause of Gender Bias for Chest X-Ray Diagnosis?},
  year      = {2023},
  pages     = {142--152},
  publisher = {Springer Nature Switzerland},
}

@InProceedings{Pleiss2017,
  author    = {Pleiss, Geoff and Raghavan, Manish and Wu, Felix and Kleinberg, Jon and Weinberger, Kilian Q},
  booktitle = {Advances in Neural Information Processing Systems},
  year      = {2017},
  title     = {On fairness and calibration},
  pages     = {5680--5689},
}

@InProceedings{Kleinberg2017,
  author    = {Kleinberg, Jon and Mullainathan, Sendhil and Raghavan, Manish},
  title     = {Inherent Trade-Offs in the Fair Determination of Risk Scores},
  year      = {2017},
  pages     = {43:1--43:23},
  publisher = {Schloss Dagstuhl – Leibniz-Zentrum für Informatik},
  volume    = {67},
  journal   = {LIPIcs, Volume 67, ITCS 2017},
}

@InProceedings{JimenezSanchez2023,
  author    = {Jiménez-Sánchez, Amelia and Juodelyte, Dovile and Chamberlain, Bethany and Cheplygina, Veronika},
  booktitle = {International Symposium on Biomedical Imaging (ISBI)},
  title     = {Detecting Shortcuts in Medical Images -- A Case Study in Chest X-rays},
  year      = {2023},
  publisher = {IEEE},
}

@Article{Irvin2019,
  author    = {Irvin, Jeremy and others},
  journal   = {Proceedings of the AAAI Conference on Artificial Intelligence},
  title     = {{CheXpert}: A Large Chest Radiograph Dataset with Uncertainty Labels and Expert Comparison},
  year      = {2019},
  issn      = {2159-5399},
  number    = {01},
  pages     = {590--597},
  volume    = {33},
  publisher = {Association for the Advancement of Artificial Intelligence (AAAI)},
}

@Article{Petersen2023b,
  author        = {Petersen, Eike and Ferrante, Enzo and Ganz, Melanie and Feragen, Aasa},
  title         = {Are demographically invariant models and representations in medical imaging fair?},
  year          = {2023},
  archiveprefix = {arXiv},
  eprint        = {2305.01397},
  primaryclass  = {cs.LG},
  publisher     = {arXiv},
  journal = {arXiv:2305.01397}
}

@Article{Banerjee2023,
  author    = {Banerjee, Imon and others},
  journal   = {Journal of the American College of Radiology},
  title     = {“{S}hortcuts” Causing Bias in Radiology Artificial Intelligence: Causes, Evaluation, and Mitigation},
  year      = {2023},
  issn      = {1546-1440},
  number    = {9},
  pages     = {842--851},
  volume    = {20},
  publisher = {Elsevier BV},
}

@InProceedings{Liu2019,
  title = 	 {The Implicit Fairness Criterion of Unconstrained Learning},
  author =       {Liu, Lydia T. and Simchowitz, Max and Hardt, Moritz},
  booktitle = 	 {Proceedings of the 36th International Conference on Machine Learning},
  pages = 	 {4051--4060},
  year = 	 {2019},
  volume = 	 {97},
  series = 	 {Proceedings of Machine Learning Research},
  publisher =    {PMLR},
}

@InProceedings{Hardt2016a,
  author    = {Hardt, Moritz and Price, Eric and Srebro, Nathan},
  booktitle = {Proceedings of the 30th International Conference on Neural Information Processing Systems},
  title     = {Equality of Opportunity in Supervised Learning},
  year      = {2016},
  pages     = {3323–3331},
  series    = {NIPS'16},
}

@Article{Brown2023,
  author    = {Alexander Brown and Nenad Tomasev and Jan Freyberg and Yuan Liu and Alan Karthikesalingam and Jessica Schrouff},
  journal   = {Nature Communications},
  title     = {Detecting shortcut learning for fair medical {AI} using shortcut testing},
  year      = {2023},
  number    = {1},
  volume    = {14},
  publisher = {Springer Science and Business Media {LLC}},
}

@InProceedings{Makar2022,
  author    = {Makar, Maggie and Packer, Ben and Moldovan, Dan and Blalock, Davis and Halpern, Yoni and D'Amour, Alexander},
  booktitle = {Proceedings of The 25th International Conference on Artificial Intelligence and Statistics},
  title     = {Causally motivated shortcut removal using auxiliary labels},
  year      = {2022},
  pages     = {739--766},
  publisher = {PMLR},
  series    = {Proceedings of Machine Learning Research},
  volume    = {151},
}

@InProceedings{Feragen2026,
  author    = {Feragen, Aasa and Petersen, Eike and Ganz-Benjaminsen, Melanie},
  pages     = {465--485},
  publisher = {Elsevier},
  title     = {The cause and effect of an {MR} image: Robustness and generalizability},
  year      = {2026},
  booktitle = {Machine Learning in {MRI} - From Methods to Clinical Translation},
}

@InProceedings{Anthis2023,
  author    = {Anthis, Jacy and Veitch, Victor},
  booktitle = {Advances in Neural Information Processing Systems},
  title     = {Causal Context Connects Counterfactual Fairness to Robust Prediction and Group Fairness},
  year      = {2023},
  pages     = {34122--34138},
  volume    = {36},
}

@InProceedings{Veitch2021,
  author    = {Victor Veitch and Alexander D'Amour and Steve Yadlowsky and Jacob Eisenstein},
  booktitle = {Advances in Neural Information Processing Systems},
  title     = {Counterfactual Invariance to Spurious Correlations: Why and How to Pass Stress Tests},
  year      = {2021},
}

@Article{Gichoya2022,
  author    = {Judy Wawira Gichoya and others},
  journal   = {The Lancet Digital Health},
  title     = {{AI} recognition of patient race in medical imaging: a modelling study},
  year      = {2022},
  number    = {6},
  pages     = {e406--e414},
  volume    = {4},
  publisher = {Elsevier {BV}},
}

@Article{VanCalster2016,
  author    = {Van Calster, Ben and Nieboer, Daan and Vergouwe, Yvonne and De Cock, Bavo and Pencina, Michael J. and Steyerberg, Ewout W.},
  journal   = {Journal of Clinical Epidemiology},
  title     = {A calibration hierarchy for risk models was defined: from utopia to empirical data},
  year      = {2016},
  issn      = {0895-4356},
  pages     = {167--176},
  volume    = {74},
  publisher = {Elsevier BV},
}

@InProceedings{OakdenRayner2020a,
  author    = {Luke Oakden-Rayner and Jared Dunnmon and Gustavo Carneiro and Christopher Re},
  booktitle = {Proceedings of the {ACM} Conference on Health, Inference, and Learning},
  title     = {Hidden stratification causes clinically meaningful failures in machine learning for medical imaging},
  publisher = {{ACM}},
  year      = {2020},
}

@InProceedings{Liu2021,
  author    = {Liu, Evan Z and Haghgoo, Behzad and Chen, Annie S and Raghunathan, Aditi and Koh, Pang Wei and Sagawa, Shiori and Liang, Percy and Finn, Chelsea},
  booktitle = {Proceedings of the 38th International Conference on Machine Learning},
  title     = {Just Train Twice: Improving Group Robustness without Training Group Information},
  year      = {2021},
  pages     = {6781--6792},
  publisher = {PMLR},
  series    = {Proceedings of Machine Learning Research},
  volume    = {139},
}

@InProceedings{Long2018,
  author    = {Long, Mingsheng and Cao, Zhangjie and Wang, Jianmin and Jordan, Michael I.},
  booktitle = {Proceedings of the 32nd International Conference on Neural Information Processing Systems},
  title     = {Conditional adversarial domain adaptation},
  year      = {2018},
  pages     = {1647–1657},
  series    = {NIPS'18},
}

@Article{Yan2020a,
  author    = {Yan, Hongliang and Li, Zhetao and Wang, Qilong and Li, Peihua and Xu, Yong and Zuo, Wangmeng},
  journal   = {IEEE Transactions on Multimedia},
  title     = {Weighted and Class-Specific Maximum Mean Discrepancy for Unsupervised Domain Adaptation},
  year      = {2020},
  issn      = {1941-0077},
  number    = {9},
  pages     = {2420--2433},
  volume    = {22},
  publisher = {Institute of Electrical and Electronics Engineers (IEEE)},
}

@Article{Sellergren2025,
  author  = {Sellergren, Andrew and others},
  journal = {arXiv:2507.05201},
  title   = {{MedGemma} Technical Report},
  year    = {2025},
}

@Article{Codella2024,
  author        = {Codella, Noel C. F. and others},
  title         = {{MedImageInsight}: An Open-Source Embedding Model for General Domain Medical Imaging},
  year          = {2024},
  archiveprefix = {arXiv},
  eprint        = {2410.06542},
  primaryclass  = {eess.IV},
  publisher     = {arXiv},
  journal = {arXiv:2410.06542}
}

@InProceedings{Akiba2019,
  author     = {Akiba, Takuya and Sano, Shotaro and Yanase, Toshihiko and Ohta, Takeru and Koyama, Masanori},
  booktitle  = {Proceedings of the 25th ACM SIGKDD International Conference on Knowledge Discovery \& Data Mining},
  title      = {Optuna: A Next-generation Hyperparameter Optimization Framework},
  year       = {2019},
  pages      = {2623--2631},
  publisher  = {ACM},
  series     = {KDD ’19},
  collection = {KDD ’19},
}

@Article{Gopalan2021,
  author        = {Gopalan, Parikshit and Kalai, Adam Tauman and Reingold, Omer and Sharan, Vatsal and Wieder, Udi},
  title         = {Omnipredictors},
  year          = {2021},
  archiveprefix = {arXiv},
  copyright     = {Creative Commons Attribution 4.0 International},
  eprint        = {2109.05389},
  primaryclass  = {cs.LG},
  publisher     = {arXiv},
  journal = {arXiv:2109.05389}
}
\end{document}